\documentclass[10pt,twocolumn,letterpaper]{article}

\usepackage{iccv}
\usepackage{times}
\usepackage{epsfig}
\usepackage{graphicx}
\usepackage{mathtools}
\usepackage{amssymb}

\usepackage[inline]{enumitem}
\usepackage{multirow}
\usepackage{array, booktabs, tabularx, makecell, multirow}
\usepackage{subcaption}
\usepackage{booktabs}
\usepackage{comment}
\usepackage{caption}
\usepackage{stfloats}
\usepackage{times}
\usepackage{epsfig}
\usepackage{ctable}
\usepackage{colortbl}
\usepackage{lipsum}

\usepackage[accsupp]{axessibility}  % Improves PDF
\usepackage{xcolor}
\usepackage{stfloats}
\usepackage{subcaption}

\newcommand\blfootnote[1]{%
  \begingroup
  \renewcommand\thefootnote{}\footnote{#1}%
  \addtocounter{footnote}{-1}%
  \endgroup
}
\usepackage[pagebackref=true,breaklinks=true,letterpaper=true,colorlinks,bookmarks=false]{hyperref}

\def\iccvPaperID{7346} % *** Enter the ICCV Paper ID here

\ificcvfinal\fi

\newsavebox\CBox
\def\textBF#1{\sbox\CBox{#1}\resizebox{\wd\CBox}{\ht\CBox}{\textbf{#1}}}

\newcommand{\OURS}{NeRF-Det}

\usepackage[breaklinks=true,bookmarks=false]{hyperref}

\iccvfinalcopy % *** Uncomment this line for the final submission

\def\iccvPaperID{****} % *** Enter the ICCV Paper ID here

\ificcvfinal\fi
\begin{document}

\definecolor{Gray}{gray}{0.92}
\definecolor{darkgreen}{rgb}{0.13, 0.55, 0.13}

%%%%%%%%% TITLE
% \title{\LaTeX\ Author Guidelines for ICCV Proceedings}

% \and

% \maketitle
\ificcvfinal\thispagestyle{empty}\fi

%%%%%%%%% TITLE
\title{\OURS: Learning Geometry-Aware Volumetric Representation \\ for Multi-View 3D Object Detection}

\author{Chenfeng Xu$^1$ \hspace{0.02cm}
Bichen Wu$^2$ \hspace{0.02cm}
Ji Hou$^2$ \hspace{0.02cm}
Sam Tsai$^2$ \hspace{0.02cm}
Ruilong Li$^1$ \hspace{0.02cm}
Jialiang Wang$^2$  \hspace{0.02cm}
Wei Zhan$^1$ \hspace{0.02cm} \vspace{0.1cm} \\ 
Zijian He$^2$ \hspace{0.02cm}
Peter Vajda$^2$ \hspace{0.02cm}
Kurt Keutzer$^1$ \hspace{0.02cm}
Masayoshi Tomizuka$^1$\\
\\
$^1$University of California, Berkeley \hspace{0.2cm}
$^2$Meta
% \{xuchenfeng, ruilong\}@berkeley.edu
}

%%%%%%%%% TEASER
\twocolumn[{%
	\renewcommand\twocolumn[1][]{#1}%
	\maketitle
	\begin{center}
		\vspace{-0.25cm}
	\includegraphics[width=.90\linewidth]{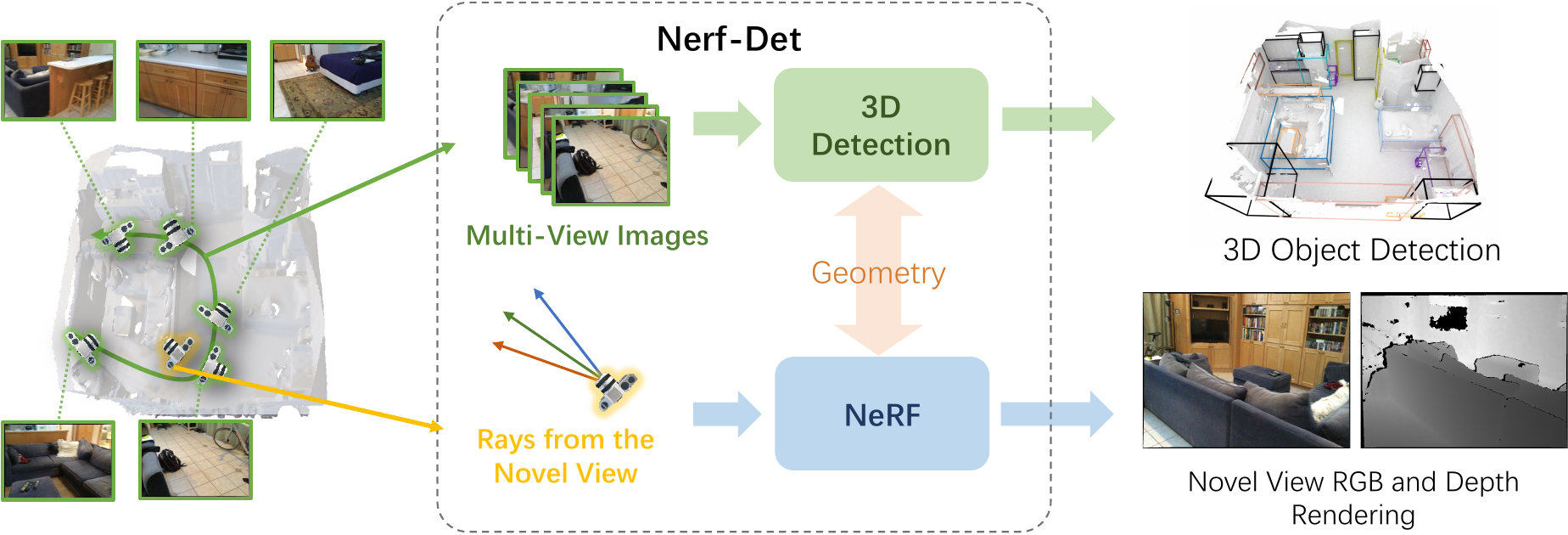}
		\captionof{figure}{
		NeRF-Det aims to detect 3D objects with only RGB images as input. To enhance detection, we propose to embed a NeRF branch with our designed synergy. The two joint branches share the same geometry representation and are trained end-to-end, which helps achieve state-of-the-art accuracy on mutli-view indoor RGB-only 3D detection, and additionally enables a generalizable novel view synthesis on new scenes without per-scene optimization.
		}
		\label{fig:teaser}
	\end{center}
}]

\maketitle
\ificcvfinal\thispagestyle{empty}\fi
\blfootnote{This work was done when Chenfeng was an intern at Meta.}
\begin{abstract}
We present \OURS{}, a novel method for indoor 3D detection with posed RGB images as input. Unlike existing indoor 3D detection methods that struggle to model scene geometry, our method makes novel use of NeRF in an end-to-end manner to explicitly estimate 3D geometry, thereby improving 3D detection performance. Specifically, to avoid the significant extra latency associated with per-scene optimization of NeRF, we introduce sufficient geometry priors to enhance the generalizability of NeRF-MLP. Furthermore, we subtly connect the detection and NeRF branches through a shared MLP, enabling an efficient adaptation of NeRF to detection and yielding geometry-aware volumetric representations for 3D detection. Our method outperforms state-of-the-arts by \textit{3.9} mAP and \textit{3.1} mAP on the ScanNet and ARKITScenes benchmarks, respectively. We provide extensive analysis to shed light on how NeRF-Det works. As a result of our joint-training design, NeRF-Det is able to generalize well to unseen scenes for object detection, view synthesis, and depth estimation tasks without requiring per-scene optimization. Code is available at \url{https://github.com/facebookresearch/NeRF-Det}.
\vspace{-0.6cm}
\end{abstract}
\section{Introduction}
\label{sec:intro}
In this paper, we focus on the task of indoor 3D object detection using posed RGB images. 3D object detection is a fundamental task for many computer vision applications such as robotics and AR/VR. The algorithm design depends on input sensors. In the past few years, most 3D detection works focus on both RGB images and depth measurements (depth images, point-clouds, \etc). While depth sensors are widely adopted in applications such as autonomous driving, they are not readily available in most AR/VR headsets and mobile phones due to cost, power dissipation, and form factor constraints. Excluding depth input, however, makes 3D object detection significantly more challenging, since we need to understand not only the semantics, but also the underlying scene geometry from RGB-only images. 

To mitigate the absence of geometry, one straightforward solution is to estimate depth. However, depth estimation itself is a challenging and open problem. For example, most monocular depth-estimation algorithms cannot provide accurate metric depth or multi-view consistency \cite{hoiem2005automatic,saxena2008make3d,karsch2014depth,Ranftl2022}. Multi-view depth-estimation algorithms can only estimate reliable depth in textured and non-occluded regions~\cite{CGV-052, seitz2006comparison}. 

Alternatively, ImVoxelNet \cite{rukhovich2022imvoxelnet} models the scene geometry implicitly by extracting features from 2D images and projecting them to build a 3D volume representation. However, such a geometric representation is intrinsically ambiguous and leads to inaccurate detection.

On the other hand, Neural Radiance Field (NeRF) \cite{mildenhall2021nerf,chen2021mvsnerf,chen2021mvsnerf} has been proven to be a powerful representation for geometry modeling. However, incorporating NeRF into the 3D detection pipeline is a complex undertaking for several reasons:
\begin{enumerate*}[label=(\roman*)]
\item Rendering a NeRF requires high-frequency sampling of the space to avoid aliasing issues \cite{mildenhall2021nerf}, which is challenging in the 3D detection pipeline due to limited resolution volume.
\item Traditional NeRFs are optimized on a per-scene basis, which is incompatible with our objective of image-based 3D detection due to the considerable latency involved.
\item NeRF makes full use of multi-view consistency to better learn geometry during training. However, a simple stitch of first-NeRF-then-perception \cite{vora2021nesf,hu2022nerf,jeong2022perfception} (i.e., reconstruction-then-detection) does not bring the advantage of multi-view consistency to the detection pipeline.
\end{enumerate*}

To mitigate the issue of ambiguous scene geometry, we propose \OURS{} to explicitly model scene geometry as an opacity field by jointly training a NeRF branch with the 3D detection pipeline. Specifically, we draw inspirations from \cite{wang2021ibrnet,yu2020pixelnerf} to project ray samples onto the image plane and extract features from the high-resolution image feature map, rather than from the low-resolution volumes, thereby overcoming the need for high-resolution volumes. To further enhance the generalizability of NeRF model to unseen scenes, we augment the image features with more priors as the input to the NeRF MLP, which leads to more distinguishable features for NeRF modeling. Unlike previous works that build a simple stitch of NeRF-then-perception, we connect the NeRF branch with the detection branch through a \textit{shared} MLP that predicts a density field, subtly allowing the gradient of NeRF branches to back-propagate to the image features and benefit the detection branch during training. We then take advantage of the uniform distribution of the volume and transform the density field into an opacity field and multiply it with the volume features. This reduces the weights of empty space in the volume feature. Then, the geometry-aware volume features are fed to the detection head for 3D bounding box regression. It is worth noting that during inference, the NeRF branch is removed, which minimizes the additional overhead to the original detector.

Our experiments show that by explicitly modeling the geometry as an opacity field, we can build a much better volume representation and thereby significantly improve 3D detection performance. Without using depth measurements for training, we improve the state-of-the-art by $3.9$ and $3.1$ mAP on the ScanNet and the ARKITScenes datasets, respectively. 
Optionally, if depth measurements are also available for training, we can further leverage depth to improve the performance, while our inference model still does not require depth sensors.
Finally, although novel-view synthesis and depth estimation are not our focus, our analysis reveals that our method can synthesize reasonable novel-view images and perform depth prediction without per-scene optimization, which validates that our 3D volume features can better represent scene geometry.

\section{Related Work}

\noindent\textbf{3D Detection in Indoor Scene.} 
%3D detection develops various methods depending on its inputs. 
3D detection utilizes various methods depending on inputs, and has achieved great success on point cloud~\cite{nie2020rfd, imvotenet, votenet, qi2018frustum, zhang2020h3dnet} and voxel representations~\cite{yi2018gspn, hou2020revealnet, hou20193d,engelmann20203d, hou2021exploring}. 3D-SIS~\cite{hou20193d} uses anchors to predict 3D bounding boxes from voxels fused by color and geometric features. The widely-used VoteNet~\cite{qi2019deep} proposes hough voting to regresses bounding box parameters from point features. 
However, depth sensors are not always readily available on many devices due to its huge power consumption, such as on VR/AR headsets. 
To get rid of sensor depth, Panoptic3D~\cite{dahnert2021panoptic} operates on point clouds extracted from predicted depth. Cube R-CNN~\cite{brazil2022omni3d} directly regresses 3D bounding boxes from a single 2D image. 

Comparably, the multi-view approach is also not limited by depth sensors and is more accurate. However, the current state-of-the-art multi-view method~\cite{rukhovich2022imvoxelnet} fuses the image naively by duplicating pixel features along the ray, which does not incorporate a sufficient amount of geometric clues. To address this, we leverage NeRF to embed geometry into the volume for better 3D detection.

\noindent\textbf{3D Reconstruction with Neural Radiance Field.}
Neural Radiance Field (NeRF)~\cite{mildenhall2021nerf} is a groundbreaking 3D scene representation that emerged three years ago, and has proven to be powerful for reconstructing 3D geometry from multi-view images~\cite{mildenhall2021nerf,barron2021mip,yariv2021volume,zhao2022human,wang2022neuris}. Early works~\cite{mildenhall2021nerf,barron2021mip,li2022tava,peng2021animatable,yu2021plenoctrees} directly optimize for per-scene density fields using differentiable volumetric rendering~\cite{max1995optical}. 
Later, NeuS~\cite{wang2021neus} and VolSDF~\cite{yariv2021volume} improve the reconstruction quality by using SDF as a replacement of density as the geometry representation. Recently, Ref-NeRF~\cite{verbin2022ref} proposes to use reflective direction for better geometry of glossy objects. Aside from aforementioned per-scene optimization methods, there are also works that aim to learn a generalizable NeRF from multiple scenes, such as IBRNet~\cite{wang2021ibrnet} and MVS-NeRF~\cite{chen2021mvsnerf}, which predict the density and color at each 3D location conditioned on the pixel-aligned image features. Despite this amazing progress, all of these methods only focus on a single task -- either novel-view synthesis or surface reconstruction. In contrast, we propose a novel method that incorporates NeRF seamlessly to improve 3D detection.

\noindent\textbf{NeRF for Perception.} Being able to capture accurate geometry, NeRF has gradually been incorporated into perception tasks such as classification \cite{jeong2022perfception}, segmentation \cite{vora2021nesf,zhi2021place}, detection \cite{hu2022nerf}, instance segmentation \cite{KunduCVPR2022PNF}, and panoptic segmentation \cite{fu2022panoptic}. However, most of them \cite{jeong2022perfception,hu2022nerf,vora2021nesf} follow the pipeline of first-NeRF-then-perception, which not only creates extra cost for perception tasks but also does not sufficiently use volumetric renderings to benefit perception during training. Besides, \cite{zhi2021place,fu2022panoptic} demonstrate that NeRF can significantly improve label efficiency for semantic segmentation by ensuring multi-view consistency and geometry constraints. Our proposed NeRF-Det method incorporates NeRF to ensure multi-view consistency for 3D detection. Through joint end-to-end training for NeRF and detection, no extra overheads are introduced during inference.

\vspace{-0.8cm}
\section{Method}
\begin{figure*}[htb!]
\centering
	\begin{center}
		\vspace{-0.5cm}
	\includegraphics[width=.93\linewidth]{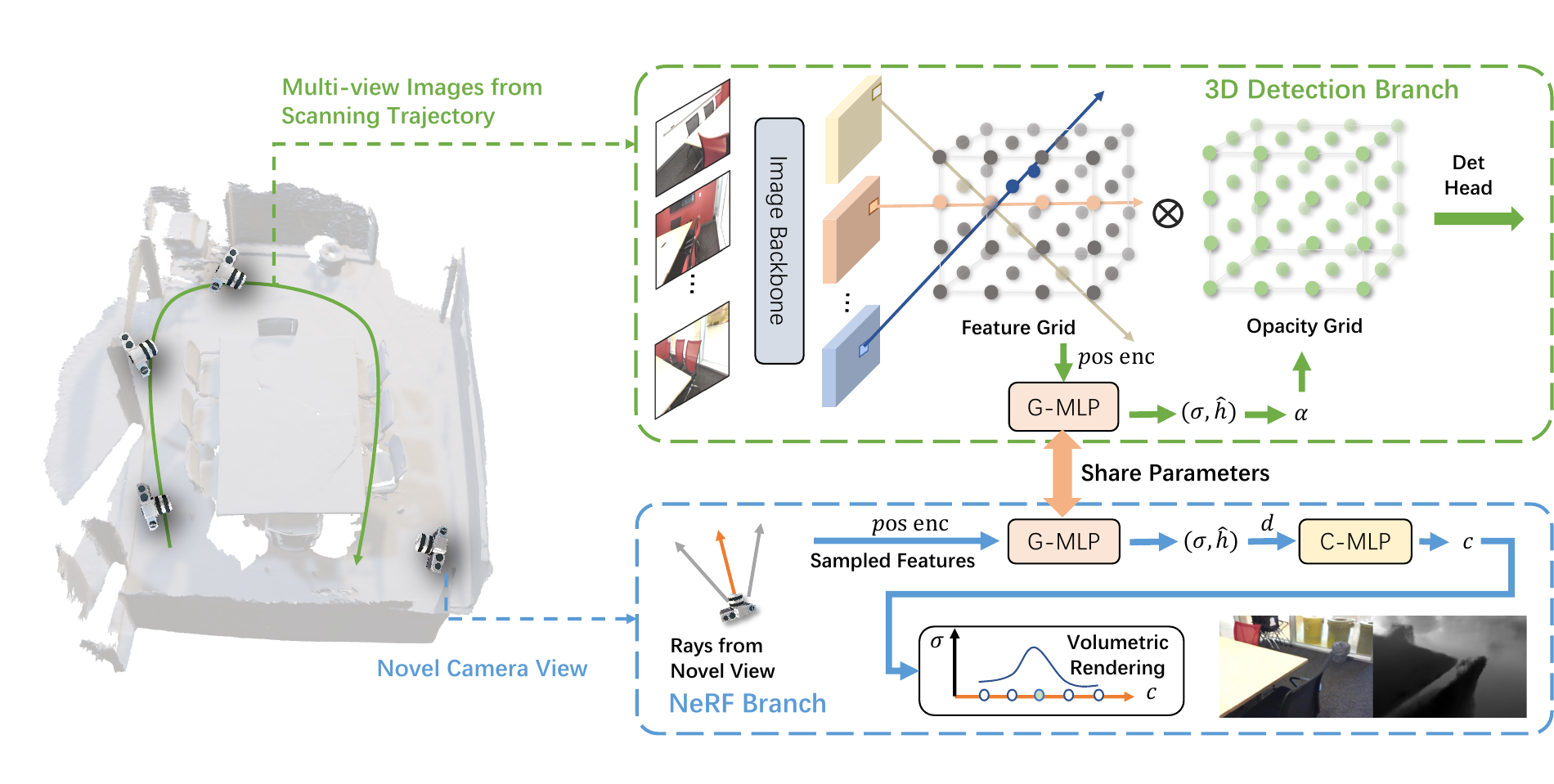}
\vspace{-0.55cm}
		\captionof{figure}{
		\textbf{The framework of NeRF-Det.} Our method leverages NeRF to learn scene geometry by estimating opacity grids. 
		With the shared geometry-MLP (G-MLP), the detection branch can benefit from NeRF in estimating opacity fields and is thus able to mask out free space and mitigate the ambiguity of the feature volume.
		}
		\label{fig:framework}
	\end{center}
\vspace{-0.65cm}
\end{figure*}

Our method, termed as \OURS{}, uses posed RGB images for indoor 3D object detection by extracting image features and projecting them into a 3D volume. We leverage NeRF to help infer scene geometry from 2D observations. To achieve this, we entangle the 3D object detection and NeRF with a \textit{shared} MLP, with which the multi-view constraint in NeRF can enhance the geometry estimation for the detection, as shown in Fig.~\ref{fig:framework}.

\subsection{3D Detection Branch}
\label{detection_branch}
In the 3D detection branch, we input posed RGB frames to the 2D image backbone, denoting the images as $I_i \in \mathbf{R}^{H_i \times W_i \times 3}$ and the corresponding intrinsic matrix and extrinsic matrix as $K \in \mathbf{R}^{3 \times 3}$ and $R_i \in \mathbf{R}^{3 \times 4}$, where $i = 1, 2, 3, ..., T$ and $T$ is the total number of views. We follow \cite{rukhovich2022imvoxelnet}, which uses an FPN \cite{lin2017feature} to fuse multi-stage features and use high resolution features, denoted as $F_i \in \mathbf{R}^{C \times H/4 \times W/4}$, to generate a 3D feature volume.

% \vspace{-0.25cm}
% \paragraph{Generating 3D Feature Volume.} 
We create the 3D feature volume by attaching 2D features from each image to their corresponding positions in 3D. We establish a 3D coordinate system, with the z-axis denoting height, and the x- and y-axis denoting two horizontal dimensions. Then, we build a 3D grid of with $N_x \times N_y \times N_z$ voxels. For each voxel center with coordinate $\mathbf{p} = (x, y, z)^T$, we project it to view-$i$ to obtain the 2D coordinates as
% \begin{equation}
%     \begin{pmatrix} u'_i \\ v'_i \\ d_i \end{pmatrix} = K' \times R_i \times \begin{pmatrix} x \\ y \\ z \\ 1 \end{pmatrix},
% \end{equation}
\begin{equation}
\vspace{-0.1cm}
    \begin{pmatrix} u'_i, v'_i, d_i \end{pmatrix}^T = K' \times R_i \times \begin{pmatrix} \mathbf{p}, 1 \end{pmatrix}^T,
\end{equation}
where $(u_i, v_i) = (u'_i/d_i, v'_i/d_i)$ is to the pixel coordinate of $\mathbf{p}$ in view-$i$. $K'$ is the scaled intrinsic matrix, considering the feature map downsampling. After building this correspondence, we assign 3D features as 
\begin{eqnarray}
    V_i(p) = \texttt{interpolate}((u_i, v_i), F_i),
\end{eqnarray}
where $\texttt{interpolate}()$ looks up the feature in $F_i$ at location $(u_i, v_i)$. Here we use nearest neighbor interpolation. For $\mathbf{p}$ that are projected outside the boundary of the image, or behind the image plane, we discard the feature and set $V_i(\mathbf{p}) = \mathbf{0}$. Intuitively, this is equivalent to shooting a ray from the origin of camera-$i$ through pixel $(u_i, v_i)$, and for all voxels that are on the ray, we scatter image features to the voxel in $V_i$. Next, we aggregate multi-view features by simply averaging all \textit{effective} features as done in ImVoxelNet \cite{rukhovich2022imvoxelnet}. Letting $M_p$ denote the number of effective 2D projections, we compute $V^{avg}(p) = \sum_{i=1}^{M_p} V_i(p)/M_p$.  

However, volume features generated this way ``over populate'' projection rays without considering empty spaces and other geometric constraints. 
% As shown in the detection branch of Fig.~\ref{fig:framework}, empty spaces in the volume are also filled with image features. 
This makes the 3D representation ambiguous and causes detection errors. To mitigate this issue, we propose to incorporate a NeRF branch to improve learning geometry for the detection branch.

\begin{figure}[t!]
\centering
\includegraphics[width=0.45\textwidth]{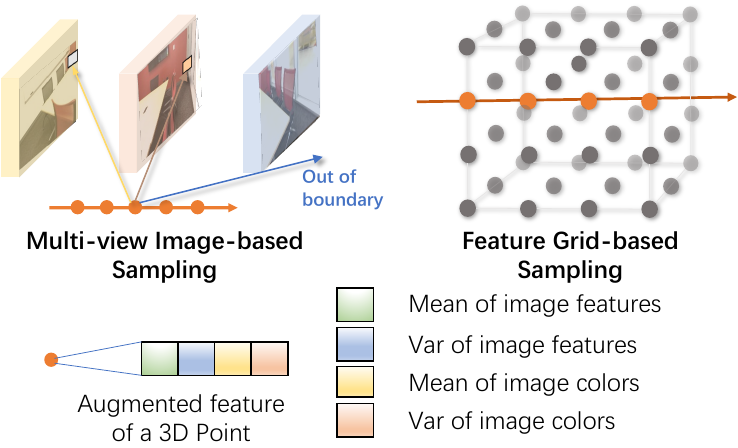}
\vspace{-0.2cm}
\caption{Different feature sampling strategies. Give a ray from a novel view, we can project the 3D points on the ray to the multi-view image features, as presented in the left part, and attach their mean/variance as well as the corresponding RGB to the 3D point. On the other hand, we can also sample features from feature grids, as shown in the right part.}
\label{fig:feature_sampling}
\vspace{-0.35cm}
\end{figure}

\subsection{NeRF Branch}
\label{sec:nerf_branch}

\paragraph{Feature Sampling.} NeRF \cite{mildenhall2021nerf} is a neural rendering method for novel view synthesis. 
However, previous NeRFs sample features from the 3D volume with high resolution, such as $128^3$ \cite{mildenhall2021nerf}. In 3D detection, we use a lower grid resolution of $40 \times 40 \times 16$, which suffers from the aliasing issue and results in degradation of geometry learning. To mitigate this, we draw inspiration from \cite{wang2021ibrnet,yu2020pixelnerf} and sample features from higher resolution 2D image feature maps, typically of size $160 \times 120$, as shown in Fig.~\ref{fig:feature_sampling}.
Specifically, we first sample points along the ray originated from the camera, i.e., $\mathbf{r}(t) = \mathbf{o} + t \times \mathbf{d}$ where $\mathbf{o}$ is the ray origin and $\mathbf{d}$ is the view direction. For a coordinate $\mathbf{p}$ sampled on the ray and a viewing direction $\mathbf{d}$, we can compute the color $\mathbf{c}(\mathbf{p},\mathbf{d})$ and density $\sigma(\mathbf{p})$ as:
\vspace{-0.3cm}
\begin{align}
\label{nerf_mlps}
    \sigma(\mathbf{p}), \mathbf{\hat{h}}(\mathbf{p})&=\textit{G-MLP}(\Bar{V}(\mathbf{p}), \gamma(\mathbf{p})), \\
    \mathbf{c}(\mathbf{p}, \mathbf{d})&=\textit{C-MLP}(\mathbf{\hat{h}}(\mathbf{p}), \mathbf{d}).
    \vspace{-0.75cm}
\end{align}
$\Bar{V}(\mathbf{p})$ is ray features aggregated and \textit{augmented} from multi-view features, and $\gamma(\mathbf{p})$ is the position encoding same as \cite{mildenhall2021nerf} while $\mathbf{\hat{h}}$ is latent features. The first MLP is termed \textit{G-MLP} for estimating geometry and the second MLP is termed \textit{C-MLP} for estimating color. For activations, we follow \cite{mildenhall2021nerf} and use ReLU for the density $\sigma(\mathbf{p})$ and sigmoid for the color $c(\mathbf{p},\mathbf{d})$.

\paragraph{Augmenting Features.}
\vspace{-0.5cm}
Although it is similar to \cite{wang2021ibrnet,yu2020pixelnerf} that use image features, it is still difficult to make \textit{G-MLP} estimate accurate geometry across different scenes by simply averaging features from multi-view features as detection branch does. Therefore, we propose to incorporate more priors to help optimize \textit{G-MLP}. Beyond averaging the projected features, we augment the sampled features with the variance from multiple views $V^{var}(p) = \sum_{i=1}^{M_p} (V_i(p) - V^{avg}(p))^2 / M_p$. The variance of the color features is able to roughly describe the occupancy of the 3D field, which has been widely used as cost volume in multi-view stereo depth estimation \cite{yang2020cost}. If the 3D location $\mathbf{p}$ is occupied, the variance of the observed features should be low under the assumption that the scene only contains Lambertian surfaces. On the other hand, if the location is in free space, different appearances would be observed from different views, and therefore the variance of color features would be high. Compared to naive average of features, variance provides a better prior for estimating scene geometry. 

In addition to extracted deep features, which are trained to be invariant to subtle appearance changes, we also augment pixel RGB values into sampled features on the ray. This is inspired by IBRNet \cite{wang2021ibrnet} where they attach the pixel RGB to the input to the MLP for better appearance modeling. We compute the mean and variance for pixel RGB values in the same way as for deep features. 
In all, the augmented feature $\Bar{V}$ is represented as a concatenation of $\{V^{avg}, V^{var}, RGB^{avg}, RGB^{var}\}$, as shown in Fig.~\ref{fig:feature_sampling}. The sampled augmented features are passed into NeRF MLPs (Eq. \ref{nerf_mlps}) to generate the density and color. We use volumetric rendering to produce the final pixel color and depth, 
\vspace{-0.1cm}
\begin{equation}
\label{equ:nerf}
    \mathbf{\hat{C}}(r) = \sum_{i=1}^{N_p} T_i\alpha_i \mathbf{c}_i, \\ D(r) = \sum_{i=1}^{N_p} T_i\alpha_i t_i,
\end{equation}
where $T_i=\exp{(-\sum_{j=1}^{i-1}\sigma_j\delta_t})$, $\alpha_i=1-\exp{(-\sigma_i\delta_t)}$, $t_i$ is the distance between sampled $i$th point between the camera, $\delta_t$ is the distance between sampled points of rays.
% In this work, we only consider using uniform sampling along the ray.

% \vspace{-0.35cm}

% \vspace{-0.5cm}
\subsection{Estimating Scene Geometry} We use an opacity field to model the scene geometry. The opacity field is a volumetric representation that reflects the probability of an object's presence in a particular area, i.e., if there is an object that cannot be seen through, the opacity field would be $1.0$ in that area. To generate the opacity field, we follow the same process of augmenting features in the detection branch as we do in the NeRF branch. 
A key ingredient to the approach is sharing the G-MLP learned in the NeRF branch with the detection branch. This enables two important capabilities. Firstly, the shared G-MLP subtly connects the two branches, allowing gradients from the NeRF branch to back-propagate and benefit the detection branch during training. Secondly, during inference, we can directly input the augmented volume features of 3D detection into the shared G-MLP since both inputs from two branches are augmented features. The output of G-MLP is the density represented as $\sigma(\mathbf{p})=\text{G-MLP}(\Bar{V}(\mathbf{p}), \gamma(\mathbf{p}))$, where $\sigma(\mathbf{p}) \in [0, \infty]$. Note that $\mathbf{p}$ is the center position of each voxel in the volume of the detection branch.

Next, we aim to transform the density field into the opacity field by $\alpha(\mathbf{p})=1-\exp{(-\sigma(\mathbf{p}) \times \delta t)}$. However, it is infeasible to calculate $\delta_t$ as we can not decide the ray direction and calculate the point distance within the undirected volume in the detection branch. Here we subtly take advantage of the uniform distribution of the voxels in the space. Thus, the $\delta_t$ in the volumetric rendering equation can be canceled out as it is a constant. So obtaining opacity can be reduced to $\alpha(\mathbf{p})=1-\exp{(-\sigma(\mathbf{p}))}$. After generating the opacity field, we multiply it with the feature grid $V^{avg}$ for 3d detection, denoted as $\alpha(\mathbf{p}) \times V^{avg}(\mathbf{p})$. 

\subsection{3D Detection Head and Training Objectives}
Our geometry-aware volumetric representation can fit to most detection heads. For fair comparison and the simplicity, we use the same indoor detection head as ImVoxelNet \cite{rukhovich2022imvoxelnet}, in which we select 27 location candidates for each objects and we use three convolution layers to predict the categories, the locations and the centerness. 

We jointly train detection and NeRF branches end-to-end. No per-scene optimization for the NeRF branch is performed in test time. For the detection branch, we supervise training with ground truth 3D bounding boxes following ImVoxelNet \cite{rukhovich2022imvoxelnet}, which computes three losses: focal loss for classification $L_{cls}$, centerness loss $L_{cntr}$, and localization loss $L_{loc}$. For the NeRF branch, we use a photo-metric loss $L_c = ||\hat{C}(r) - \hat{C}_{gt}(r)||_2$. When depth ground truth is used, we can further supervise the expected depth of the scene geometry as $L_d = ||D(r) - D_{gt}(r)||$ where $D(r)$ is computed with Equ.~\ref{equ:nerf}. The final loss $L$ is given by
\begin{equation}
    L = L_{cls} + L_{cntr} + L_{loc} + L_c + L_d.
\end{equation}
Even though we \textit{optionally} use depth during training, it is not required in inference. Also our trained network is generalizable to new scenes which are never seen during \mbox{training}.

\section{Experimental Results}
\label{sec:results}
\begin{table*}
\centering
\captionsetup{aboveskip=2pt}\captionsetup{belowskip=0pt}\captionof{table}{3D Detection with multi-view RGB inputs on ScanNet. The first block of the table includes point-cloud based and RGBD-based methods, and the rest are multi-view RGB-only detection methods. $\dagger$ means our reproduction of ImVoxelNet \cite{rukhovich2022imvoxelnet} using the official code. * indicates the NeRF-Det with depth supervision. 1x and 2x refer that we train the model with the same and two times of the training iteration wrt. the original iterations of ImVoxelNet, respectively.}
\label{tab:main_result}
% \tiny
\resizebox{\textwidth}{!}{
\begin{tabular}{l|c|c|c|c|c|c|c|c|c|c|c|c|c|c|c|c|c|c|c}
% @{\hspace{1.0\tabcolsep}}c@{\hspace{1.0\tabcolsep}}c@{\hspace{1.0\tabcolsep}}c@{\hspace{1.0\tabcolsep}}c} 
% \begin{tabular}{ p{25mm}<{\centering}| p{10mm}<{\centering} | p{12mm}<{\centering}  p{10mm}<{\centering}  p{10mm}<{\centering} p{10mm}<{\centering} p{10mm}<{\centering} p{12mm}<{\centering} | p{10mm}<{\centering}}

\toprule
\textbf{Methods} & cab & bed & chair & sofa & tabl & door & wind & bkshf & pic & cntr & desk & curt & fridg & showr & toil & sink & bath & ofurn & \textbf{mAP@.25} \\
\midrule

Seg-Cluster \cite{wang2018sgpn} & 11.8 & 13.5 & 18.9 & 14.6 & 13.8 & 11.1 & 11.5 & 11.7 & 0.0 &  13.7 & 12.2 & 12.4 & 11.2 & 18.0 & 19.5 & 18.9 & 16.4 & 12.2 & 13.4 \\

Mask R-CNN \cite{he2017mask} & 15.7 & 15.4 & 16.4 & 16.2 & 14.9 & 12.5 & 11.6 & 11.8 & 19.5 & 13.7 & 14.4 & 14.7 & 21.6 & 18.5 & 25.0 & 24.5 & 24.5 & 16.9 & 17.1 \\

SGPN \cite{wang2018sgpn} & 20.7 & 31.5 & 31.6 & 40.6 & 31.9 & 16.6 & 15.3 & 13.6 & 0.0 & 17.4 &  14.1 & 22.2 & 0.0 & 0.0 & 72.9 & 52.4 & 0.0 & 18.6 & 22.2 \\

3D-SIS \cite{hou20193d} & 12.8 & 63.1 & 66.0 & 46.3 & 26.9 & 8.0 & 2.8 & 2.3 & 0.0 & 6.9 & 33.3 & 2.5 & 10.4 & 12.2 & 74.5 & 22.9 & 58.7 & 7.1 & 25.4 \\

3D-SIS (w/ RGB) \cite{hou20193d} & 19.8 & 69.7 & 66.2 & 71.8 & 36.1 & 30.6 & 10.9 & 27.3 & 0.0 & 10.0 & 46.9 & 14.1 & 53.8 & 36.0 & 87.6 & 43.0 & 84.3 & 16.2 & 40.2 \\

VoteNet \cite{qi2019deep} & 36.3 & 87.9 & 88.7 & 89.6 & 58.8 & 47.3 & 38.1 & 44.6 & 7.8 & 56.1 & 71.7 & 47.2 & 45.4 & 57.1 & 94.9 & 54.7 & 92.1 & 37.2 & 58.7 \\

FCAF3D \cite{rukhovich2022fcaf3d} & 57.2 & 87.0 & 95.0 & 92.3 & 70.3 & 61.1 & 60.2 & 64.5 & 29.9 & 64.3 & 71.5 & 60.1 & 52.4 & 83.9 & 99.9 & 84.7 & 86.6 & 65.4 & 71.5 \\

CAGroup3D \cite{wang2022cagroupd} & 60.4 & 93.0 & 95.3 & 92.3 & 69.9 & 67.9 & 63.6 & 67.3 & 40.7 & 77.0 & 83.9 & 69.4 & 65.7 & 73.0 & 100.0 & 79.7 & 87.0 & 66.1 & 75.12 \\

\midrule
ImVoxelNet-R50-1x & 28.5 & 84.4 & 73.1 & \textbf{70.1} & 51.9 & 32.2 & 15.0 & 34.2 & 1.6 & 29.7 & 66.1 & \textbf{23.5} & 57.8 & 43.2 & 92.4 & 54.1 & 74.0 &34.9& 48.1 \\

ImVoxelNet$^\dagger$-R50-1x & 31.6 & 81.8 & 74.4 & 69.3 & \textbf{53.6} & 29.7 & 12.9 & \textbf{50.0} & 1.3 & 32.6 & \textbf{69.2} & 12.7 & 54.6 & 31.8 & \textbf{93.1} & \textbf{55.8} & 68.2 & 31.8 & 47.5 \\

\rowcolor[gray]{.9} 
NeRF-Det-R50-1x & 32.7 & 82.6 & 74.3 & 67.6 & 52.3 & 34.4 & 17.3 & 40.1 & \textbf{2.0} & 49.2 & 67.4 & 20.0 & 57.2 & 41.0 & 90.9 & 52.3 & 74.0 & \textbf{33.7} & 49.5~\small{\textcolor{darkgreen}{(+2.0)}} \\ % Us!
\rowcolor[gray]{.9} 
NeRF-Det-R50-1x* & \textbf{32.7} & \textbf{84.7} & \textbf{74.6} & 62.7 & 52.7 & \textbf{35.1} & \textbf{17.7} & 48.4 & 0.0 & \textbf{49.8} & 64.6 & 18.5 & \textbf{60.3} & \textbf{48.3} & 90.7 & 51.0 & \textbf{76.8} & 30.4 & \textBF{50.1}~\small{\textcolor{darkgreen}{(+2.6)}} \\  \midrule % Us! 

ImVoxelNet$^\dagger$-R50-2x & 34.5 & 83.6 & 72.6 & 71.6 & 54.2 & 30.3 & 14.8 & 42.6 & \textbf{0.8} & 40.8 & 65.3 & 18.3 & 52.2 & 40.9 & 90.4 & \textbf{53.3} & 74.9 & 33.1 & 48.4 \\ 

\rowcolor[gray]{.9} 
NeRF-Det-R50-2x & 37.2 & \textbf{84.8} & \textbf{75.0} & \textbf{75.6} & 51.4 & 31.8 & \textbf{20.0} & 40.3 & 0.1 & 51.4 & 69.1 & \textbf{29.2} & \textbf{58.1} & \textbf{61.4} & 91.5 & 47.8 & \textbf{75.1} & 33.6 & \textbf{52.0}~\small{\textcolor{darkgreen}{(+3.6)}} \\ % Us!
\rowcolor[gray]{.9} 
NeRF-Det-R50-2x* & \textbf{37.7} & 84.1 & 74.5 & 71.8 & \textbf{54.2} & \textbf{34.2}  & 17.4 & \textbf{51.6} & 0.1 & \textbf{54.2} & \textbf{71.3} & 16.7 & 54.5 & 55.0 & \textbf{92.1} & 50.7 & 73.8 & \textbf{34.1} & 51.8~\small{\textcolor{darkgreen}{(+3.4)}} \\ % Us!
\midrule
ImVoxelNet$^\dagger$-R101-2x & 30.9 & 84.0 & \textbf{77.5} & 73.3 & 56.7 & 35.1 & \textbf{18.6} & 47.5 & 0.0 & 44.4 & 65.5 & 19.6 & 58.2 & 32.8 & 92.3 & 40.1 & 77.6 & 28.0 & 49.0 \\
\rowcolor[gray]{.9} 
NeRF-Det-R101-2x  & 36.8 & \textbf{85.0} & 77.0 & 73.5 & 56.9 & \textbf{36.7} & 14.3 & \textbf{48.1} & 0.8 & \textbf{49.7} & \textbf{68.3} & 23.5 & 54.0 & \textbf{60.0} & 96.5 & 49.3 & 78.4 & 38.4  & 52.9~\small{\textcolor{darkgreen}{(+3.9)}}  \\
\rowcolor[gray]{.9} 
NeRF-Det-R101-2x* & \textbf{37.6} & 84.9 & 76.2 & \textbf{76.7} & \textbf{57.5} & 36.4 & 17.8 & 47.0 & \textbf{2.5} & 49.2 & 52.0 & \textbf{29.2} & \textbf{68.2} & 49.3& \textbf{97.1} & \textbf{57.6} & \textbf{83.6} & \textbf{35.9} & \textBF{53.3}~\small{\textcolor{darkgreen}{(+4.3)}}  \\ % Us!
\bottomrule
\end{tabular}}
%\vspace{-0.35cm}
\end{table*}

\begin{table*}
\centering
\captionsetup{aboveskip=2pt}\captionsetup{belowskip=0pt}\captionof{table}{Comparison experiments "whole-scene" of ARKITScenes validation set.}
\label{tab:main_result_arkit}
% \tiny
\resizebox{\textwidth}{!}{
\begin{tabular}{l|c|c|c|c|c|c|c|c|c|c|c|c|c|c|c|c|c|c}
% @{\hspace{1.0\tabcolsep}}c@{\hspace{1.0\tabcolsep}}c@{\hspace{1.0\tabcolsep}}c@{\hspace{1.0\tabcolsep}}c} 
% \begin{tabular}{ p{25mm}<{\centering}| p{10mm}<{\centering} | p{12mm}<{\centering}  p{10mm}<{\centering}  p{10mm}<{\centering} p{10mm}<{\centering} p{10mm}<{\centering} p{12mm}<{\centering} | p{10mm}<{\centering}}

\toprule
\textbf{Methods} & cab & fridg & shlf & stove & bed & sink & wshr & tolt & bthtb & oven & dshwshr & frplce & stool & chr & tble & TV & sofa & \textbf{mAP@.25} \\
\midrule

ImVoxelNet-R50 & 32.2& 34.3 & 4.2 & 0.0 & 64.7 & 20.5 & 15.8 & 68.9 & 80.4 & 9.9 & 4.1 & 10.2 & \textbf{0.4} & \textbf{5.2} & 11.6 & 3.1 & 35.6 & 23.6\\

\rowcolor[gray]{.9} 
NeRF-Det-R50 & \textbf{36.1} & \textbf{40.7} & \textbf{4.9} & \textbf{0.0} & \textbf{69.3} & \textbf{24.4} & \textbf{17.3} & \textbf{75.1} & \textbf{84.6} & \textbf{14.0} & \textbf{7.4} & \textbf{10.9} & 0.2 & 4.0 & \textbf{14.2} & \textbf{5.3} & \textbf{44.0} & \textBF{26.7}~\small{\textcolor{darkgreen}{(+3.1)}}\\ % Us!

\bottomrule
\end{tabular}}
% \vspace{-0.5cm}
\end{table*}

\noindent\textbf{Implementation details.} Our detection branch mainly follows ImVoxelNet, including backbones, detection head, resolutions and training recipe \textit{etc}. Please refer to supplemental material for more details.

Our implementation is based on MMDetection3D~\cite{mmdet3d2020}. To the best of our knowledge, we are the first to implement NeRF in MMDetection3D. We are also the first to conduct NeRF-style novel view synthesis and depth estimation on the whole ScanNet dataset, while prior works only test on a small number of scenes \cite{xu2022point,wei2021nerfingmvs}. The code will be released for future research.

\subsection{Main results}
\begin{table}
\centering
\captionsetup{aboveskip=2pt}\captionsetup{belowskip=0pt}\captionof{table}{Ablation on scene geometry modelings. GT-Depth indicates ground truth depth for placing 2D features in 3D volume. NeuralRecon-Depth indicates NeuralRecon~\cite{sun2021neucon} pre-trained on ScanNetV2 is used to predict the depth. Depth predictions are used in both training and inference.}
\label{tab:geometry-evaluation}
\footnotesize
\resizebox{.45\textwidth}{!}{
\begin{tabular}{l|c|c}
% @{\hspace{1.0\tabcolsep}}c@{\hspace{1.0\tabcolsep}}c@{\hspace{1.0\tabcolsep}}c@{\hspace{1.0\tabcolsep}}c} 
% \begin{tabular}{ p{25mm}<{\centering}| p{10mm}<{\centering} | p{12mm}<{\centering}  p{10mm}<{\centering}  p{10mm}<{\centering} p{10mm}<{\centering} p{10mm}<{\centering} p{12mm}<{\centering} | p{10mm}<{\centering}}

\toprule
\textbf{Methods} & mAP@.25 & mAP@.50 \\
\midrule
% ImVoxelNet-R50-1x (baseline) & 47.5 & 23.3 \\
% GT-Depth-R50-1x (upper-bound) & 51.9 \textcolor{darkgreen}{(+4.4)} & 27.0 \textcolor{darkgreen}{(+3.7)}\\
% NeuralRecon-Depth-R50-1x & 47.8 \textcolor{darkgreen}{(+0.4)} & 20.1 \textcolor{red}{(-3.2)}\\
% Cost-Volume-R50-1x (sigmoid) & 48.5 \textcolor{darkgreen}{(+1.0)} & 23.6 \textcolor{darkgreen}{(+0.3)} \\
% \rowcolor[gray]{.9} 
% NeRF-Det-R50-1x (w/o NeRF branch) & 48.2 \textcolor{darkgreen}{(+0.7)} & 23.6 \textcolor{darkgreen}{(+0.3)} \\
% \rowcolor[gray]{.9} 
% NeRF-Det-R50-1x & 49.5 \textcolor{darkgreen}{(+2.0)} & 24.1 \textcolor{darkgreen}{(+0.8)} \\ % Us!
% \rowcolor[gray]{.9} 
% NeRF-Det-R50-1x* & 50.1 \textcolor{darkgreen}{(+2.6)} & 24.4 \textcolor{darkgreen}{(+1.1)}\\
% \midrule
ImVoxelNet-R50-2x (baseline) & 48.4 & 23.7 \\
GT-Depth-R50-2x (upper-bound) & 54.5 \textcolor{darkgreen}{(+6.1)} & 28.2 \textcolor{darkgreen}{(+4.5)} \\
NeuralRecon-Depth-R50-2x & 48.8 \textcolor{darkgreen}{(+0.4)} & 21.4 \textcolor{red}{(-2.3)}\\
Cost-Volume-R50-2x (sigmoid) & 49.3 \textcolor{darkgreen}{(+0.9)} & 24.4 \textcolor{darkgreen}{(+0.7)} \\
NeRF-Det-R50-2x (w/o NeRF) & 49.2 \textcolor{darkgreen}{(+0.8)} & 24.6 \textcolor{darkgreen}{(+0.9)} \\
\rowcolor[gray]{.9} 
NeRF-Det-R50-2x & 52.0 \textcolor{darkgreen}{(+3.6)} & 26.1 \textcolor{darkgreen}{(+2.5)} \\ % Us!
\rowcolor[gray]{.9} 
NeRF-Det-R50-2x* & 51.8 \textcolor{darkgreen}{(+3.4)} & 27.4 \textcolor{darkgreen}{(+3.7)} \\

% ImVoxelNet (1x) & 47.5 & 23.3 \\
% ImVoxelNet (2x) & 48.4 & 24.4 \\
% ImVoxelNet (4x) & 48.9 & 25.1 \\
% ImVoxelNet (1x w/ depth from NeuralRecon) & 47.8 & 20.1\\
% % ImVoxelNet (1x w/ depth from GT) & 51.9 & 27.0 \\
% \rowcolor[gray]{.9} 
% NeRF-Det (1x) & 49.4 & 24.1 \\ % Us!
% \rowcolor[gray]{.9} 
% % NeRF-Det (2x)  & 52.0 & 26.1 \\ % Us!
% % \rowcolor[gray]{.9} 
% % NeRF-Det (4x) & 51.6  & 27.5 \\ % Us!
% \rowcolor[gray]{.9} 
% NeRF-Det (1x w/ depth supervision) & 50.1 & 24.4 \\ % Us!

\bottomrule
\end{tabular}}
\vspace{-0.55cm}
\end{table}
\paragraph{Quantitative results.}
We compare \OURS{} with point-cloud based methods~\cite{wang2018sgpn,hou20193d,qi2019deep}, RGB-D based methods~\cite{hou20193d,he2017mask} and the state-of-the-art RGB-only method ImVoxelNet~\cite{rukhovich2022imvoxelnet} on ScanNet, as shown in Tab.~\ref{tab:main_result}. 

With ResNet50 as the image backbone, we observe that \texttt{NeRF-Det-R50-1x} outperforms \texttt{ImVoxelNet-R50-1x} by $2.0$ mAP. On top of that, NeRF-Det with depth supervision, denoted as \texttt{NeRF-Det-R50-1x*}, further improves the detection performance by $0.6$ mAP compared to only RGB supervision \texttt{NeRF-Det-R50-1x}.

We denote the total training iterations of ImVoxelNet from the official code as \texttt{1x} in the Tab.~\ref{tab:main_result}. Yet the \texttt{1x} setting only iterates through each scene roughly $36$ times, which is far from sufficient for optimizing the NeRF branch, which requires thousands of iterations to optimize one scene, as indicated in~\cite{mildenhall2021nerf,wang2021ibrnet,barron2021mip,li2022tava}. Thus, we further conduct experiments with \texttt{2x} training iterations to fully utilize the potential of NeRF, and we observe that \texttt{NeRF-Det-R50-2x} reaches $52.0$ mAP, surpassing ImVoxelNet by $3.6$ mAP under the same setting (\texttt{ImVoxelNet-R50-2x}). It is worth mentioning that we do not introduce any extra data/labels to get such an improvement. If we further use depth supervision to train the NeRF branch (\texttt{NeRF-Det-R50-2x*}), the detection branch is further improved by $1.3$ in mAP@.50 as shown in Tab.~\ref{tab:geometry-evaluation}. This validates better geometry modeling (via depth supervision) could be helpful to the 3D detection task. While NeRF-Det provides an efficient method to incorporate depth supervision during the training process, introducing depth supervision in ImVoxelNet is difficult.

Moreover, when substituting ResNet50 with ResNet101, we achieve $52.9$ mAP@.25 on ScanNet, which outperforms ImVoxelNet on the same setting over 3.9 points. With depth supervision, \texttt{NeRF-Det-R101-2x*} reduces the gap between RGB-based method ImVoxelNet~\cite{rukhovich2022imvoxelnet} and point-cloud based method VoteNet~\cite{qi2019deep} by half (from $10$ mAP  $\rightarrow{}$ $5$ mAP). Besides, we conduct experiments on the ARKitScenes (see Tab.~\ref{tab:main_result_arkit}). The 3.1 mAP improvement further demonstrates the effectiveness of our proposed method.

\begin{figure*}[t!]
	\begin{center}
% 		\vspace{-0.4cm}
	\includegraphics[width=.95\linewidth]{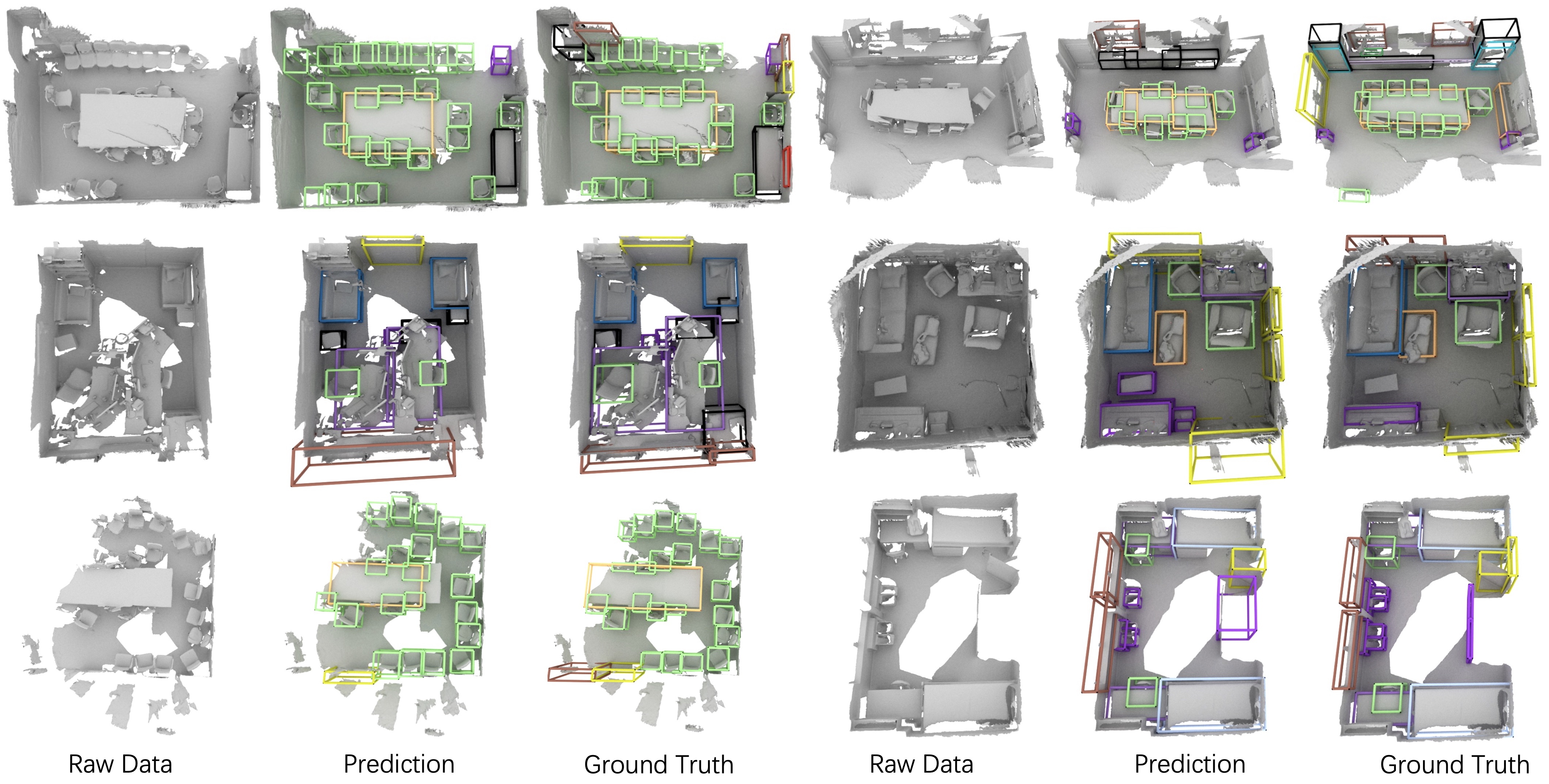}
 	% \vspace{-0.25cm}
		\captionof{figure}{
		Qualitative results of the predicted 3D bounding box on top of NeRF-Det-R50-2x. Note that our approach takes only posed RGB images as input, the reconstructed mesh is only for visualization purpose.
		}
		\label{fig:bbox_vis_results}
	\end{center}
 	\vspace{-0.35cm}
\end{figure*}

% \vspace{-0.5cm}
\paragraph{Qualitative results.}
We visualize the predicted 3D bounding boxes from \texttt{NeRF-Det-R50-2x} on scene meshes, as shown in Fig.~\ref{fig:bbox_vis_results}. We observe that the proposed method gets accurate detection prediction even on the extremely dense scenes, \textit{e.g.}, the first row and the left of the third row. The chairs are crowded in the room, and some of which are inserted into tables and are heavily occluded. Yet our method is able to detect them accurately. On the other hand, NeRF-Det can also tackle multiple scales, as shown in the second row and the right side of the third row, in which there are variant objects with difference sizes, like garbage bins, chairs, tables, doors, windows, and sofas \textit{etc}.

\vspace{-0.5cm}
\paragraph{Analysis of scene geometry modeling.}
As stated in the method section, we mitigate the ambiguity of the volume representation by learning an opacity field. Furthermore, we explore different scene geometry modeling methods, from using depth map to cost volume \cite{sun2021neucon,yang2020cost}, in Tab.~\ref{tab:geometry-evaluation}. 

\noindent\textit{Using the Depth Map.} In this experiment, we assume we have access to depth maps during \textit{both training and inference}. When building the voxel feature grid, instead of scattering the features on all points along the ray, we only place features to a single voxel cell according to the depth maps. Intuitively, this leads to less ambiguity in the volume representation, so we should observe better performance for detection. As a proof of concept, we first use ground-truth depth that comes with the dataset. This serves as an upper-bound for NeRF-Det as it provides a perfect geometry modeling. It indeed achieves a high detection accuracy of 54.5 mAP@.25 and 28.2 mAP@.50 (see second row), improving the baseline by 6.1 mAP@.25 and 4.5 mAP@.40. But in practice we cannot get access to the ground-truth depth maps. Thus, we try instead to render depth maps using out-of-box geometry reconstruction from NeuralRecon \cite{sun2021neucon}.

The results are shown in the third row of Tab.~\ref{tab:geometry-evaluation}. We observe that the depth estimation from NeuralRecon significantly degenerates the detection performance by $2.3$ mAP@.50 as compared to plain ImVoxelNet, demonstrating that the estimation error of depth maps propagates that inaccuracy through the detection pipeline.

\noindent\textit{Cost Volume.} Next, we compare our method with cost-volume based methods~\cite{yang2020cost,park2022time}. A common way to compute cost volume is using covariance~\cite{yang2020cost} between source view and reference view. This is similar to our method which calculates the variance of multiple input views. Following~\cite{yang2020cost}, we first use several 3D convolution layers to encode the cost volume, then get the probability volume via sigmoid, and multiply it with the feature volume $V^{avg}$. The results are in the fourth row of Tab.~\ref{tab:geometry-evaluation}. We can see that the cost-volume based method improves ImVoxelNet by 0.9 mAP@.25 and 0.7 mAP@.50 respectively. It is noteworthy to mention that if we remove the NeRF branch, our method is very similar to a cost-volume-based method with the differences being: 1) we augment the variance in the cost volume by mean volume and color volume, 2) we use the MLP and opacity function instead of sigmoid to model the scene geometry. The result is shown in the fifth row of Tab.~\ref{tab:geometry-evaluation}. We observe that the result is very close to a cost-volume-based method and that both ours and the cost-volume method lead to improvement over ImVoxelNet. 
% This demonstrates that cost-volume based method can benefit to detection task via better geometry modeling.

In contrast to explicitly estimating the depth and cost volume, we leverage NeRF to estimate the opacity field for the scene geometry. As shown in the gray part of Tab.~\ref{tab:geometry-evaluation}, with the opacity field modeled using NeRF, the performance is significantly improved by +3.6 mAP@.25 and +2.5 mAP@.50 compared to the baseline. After using depth supervision, NeRF-Det is able to achieve larger improvement with +3.7 mAP@.50 (the last row). As shown in Tab.~\ref{tab:geometry-evaluation}, our method of using NeRF to model scene geometry is more effective than using predicted depth or cost volume.

\begin{table}
\centering
\captionsetup{aboveskip=0pt}\captionsetup{belowskip=0pt}\captionof{table}{
The first group represents using NeRF-RPN training set, and the second group represents using ScanNet training set. The latency is measured on one V100.}
\label{tab:compare_with_nerf-then-det}
\footnotesize
\resizebox{0.5\textwidth}{!}{
\begin{tabular}{c|c|c|c}
\toprule

Method & AP25 & AP50 & Latency \\
\midrule
% NeRF-RPN-R50 (NeRF-then-det) & 28.50 & 2.00 & $\thicksim$ 10 + 7.14s  \\
NeRF-RPN-R50\cite{hu2022nerf} (NeRF-then-det) & 33.13 & 5.12 & $\thicksim$846.8s  \\
% ImVoxelNet (= NeRF-Det w/o NeRF) & 26.42 & 3.76 & 0.526s \\
% ImVoxelNet + Point-cloud supervision & 29.95 & 4.74  & 0.526s \\
\rowcolor[gray]{.9} 
NeRF-Det-R50 (joint NeRF-and-Det) & 35.13 &  7.80 & 0.554s \\
\rowcolor[gray]{.9} 
% NeRF-Det-R50 (test-time finetune) & 35.75 & 7.94 & $\thicksim$600s+0.554s \\
\midrule
\rowcolor[gray]{.9} 
NeRF-Det-R50$\dagger$ (joint NeRF-and-Det) & 61.48 & 25.45 & 0.554s \\
%\rowcolor[gray]{.9} 
% NeRF-Det-R50 (test-time finetune) & 62.10 & 26.07 & $\thicksim$600s+0.554s \\
\bottomrule
\end{tabular}}
\vspace{-0.3cm}
\end{table}

\paragraph{Comparison to NeRF-then-Det method.} We compare our proposed NeRF-Det, a joint NeRF-and-Det method, to NeRF-RPN \cite{hu2022nerf} which is a NeRF-then-Det method, as shown in Tab.~\ref{tab:compare_with_nerf-then-det}. We choose 4 scenes as validation set that are not included in both NeRF-RPN and ScanNet train set. We use the official code and provided pre-trained models of NeRF-RPN to evaluate AP. Experiments in the first group demonstrate that our proposed joint-NeRF-and-Det paradigm is more effective than first-NeRF-then-Det method NeRF-RPN, with much faster speed. 

The second group of Tab.~\ref{tab:compare_with_nerf-then-det} shows that directly using our model (NeRF-Det-R50-2x in Tab. \ref{tab:main_result}) has drastic improvements relative to NeRF-RPN. Although our model is trained on large-scale dataset, we emphasize that this is our advantages since it is hard to apply NeRF-RPN on the whole ScanNet ($\thicksim$1500 scenes) given the heavy overhead. 

\begin{figure*}[t!]
	\begin{center}
		\vspace{-0.35cm}
	\includegraphics[width=.95\linewidth]{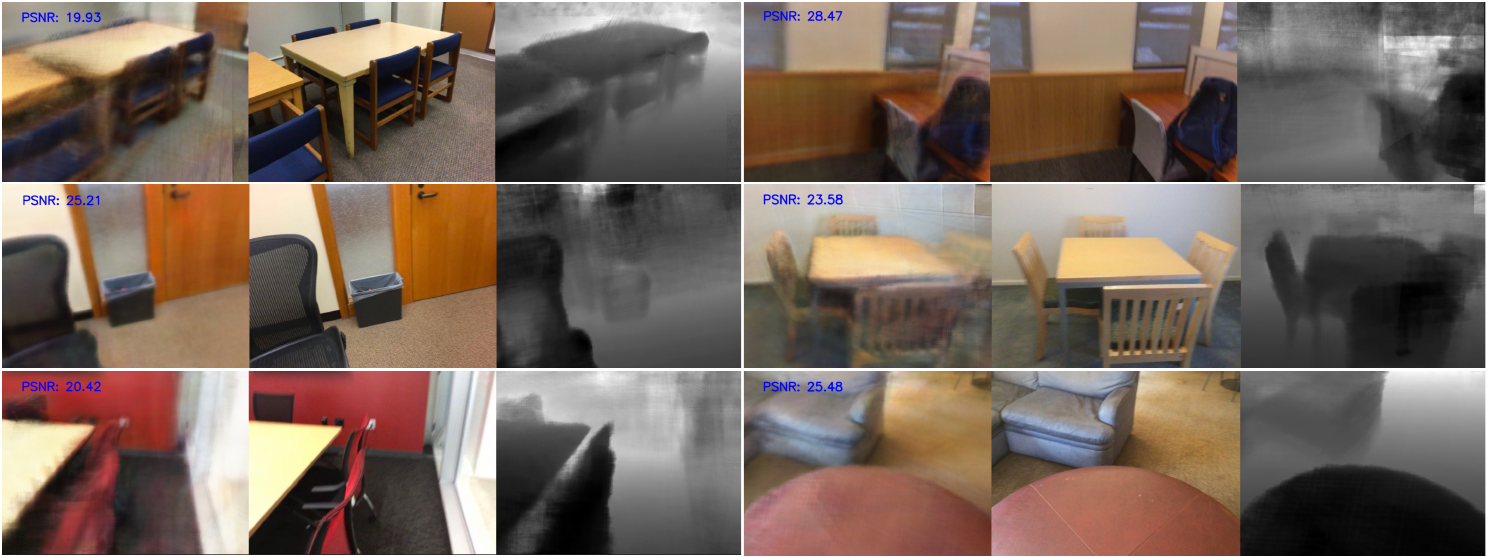}
	\vspace{-0.25cm}
		\captionof{figure}{
		Novel-view synthesis results on top of NeRF-Det-R50-2x*. For each triplet group, the left figure is the synthesized results, the middle one is the ground truth RGB image, and the right part is the estimated depth map. Note that the visualization is from test set, which is never seen during training.
		}
		\label{fig:nerf_vis}
	\end{center}
	\vspace{-0.8cm}
\end{figure*}

% \vspace{-0.55cm}
\paragraph{Is NeRF branch able to learn scene geometry?} We hypothesize that the better detection performance comes from better geometry. To verify this, we perform novel view synthesis and depth estimation from the prediction of the NeRF branch. The underlying assumption is that if our model has correctly inferred the scene geometry, we should be able to synthesize RGB and depth views from novel camera positions. We first visualize some of the synthesized images and depth estimation in Fig.~\ref{fig:nerf_vis}. The image and depth map quality look reasonable and reflects the scene geometry well.

Quantitatively, we evaluate novel-view synthesis and depth estimation by following the protocol of IBRNet \cite{wang2021ibrnet}. We select 10 novel views and randomly sample 50 nearby source views. Then, we average the evaluation results of the 10 novel views for each scene, and finally report average results on all 312 scenes in the validation set, as shown in Tab.~\ref{tab:nerf_evaluate}. Although novel-view synthesis and depth estimation are not the main focus of this paper, we achieve an average of 20+ PSNR for novel view synthesis, without per-scene training. For depth estimation, we achieves an RMSE of 0.756 on all scenes. While this performance falls behind state-of-the-art depth estimation methods, we note that \cite{wei2021nerfingmvs} reported average RMSE of 1.0125 and 1.0901 using Colmap \cite{schoenberger2016sfm} and vanilla NeRF  \cite{mildenhall2021nerf} for depth estimation on selected scenes in ScanNet. This comparison verifies that our method can render reasonable depth.

\subsection{Ablation Study}
We conduct multiple ablation studies on how different components affect the performance of NeRF-Det, including different feature sampling strategies, whether to share G-MLP, different losses and different features feed into the G-MLP. Besides, although novel-view synthesis is not our focus in this paper, we also provide some analysis for the novel-view synthesis results coming out of the NeRF branch, and show how the NeRF branch is influenced by the detection branch during joint training. All experiments in the ablation studies are based on \texttt{NeRF-Det-R50-1x*}.

\begin{table}
\centering
\captionsetup{aboveskip=2pt}\captionsetup{belowskip=0pt}\captionof{table}{Ablation study on G-MLP and different ways to sample features onto the ray in the NeRF branch.}
\label{tab:feature_sample}
\footnotesize
\resizebox{0.47\textwidth}{!}{
\begin{tabular}{c|c|c|c}
\toprule
Share G-MLP & Sample source & mAP@.25 & mAP@.50 \\
\midrule
\multirow{2}{*}{$\checkmark$}  &3D volume & 49.4 & 24.0 \\
&Multi-view 2D feature & 50.1 & 24.4\\
\midrule
\multirow{2}{*}{} &3D volume& 48.2 & 23.8 \\
&Multi-view 2D feature & 48.1 & 23.8\\

\bottomrule
\end{tabular}}
\vspace{-0.5cm}
\end{table}
\vspace{-0.55cm}
\paragraph{Ablation on G-MLP and feature sampling strategy.} As indicated in Sec.~\ref{sec:nerf_branch}, the key ingredient in our pipeline is a shared G-MLP which enables the constraint of multi-view consistency to be propagated from NeRF branch to detection branch. We conduct an ablation study as shown in Tab. \ref{tab:feature_sample}. Without shared G-MLP, the performance drops drastically from 50.1 to 48.1 at mAP@.25, shown in the fifth row of Tab. \ref{tab:geometry-evaluation}. In this case, the influence of multi-view consistency is only propagated into image backbone, significantly limiting the improvement created by NeRF.

Moreover, as mentioned in Sec.~\ref{sec:nerf_branch}, we sample point features along the ray from the multi-view image features instead of the low-resolution volume features. This ablation is shown in Tab.~\ref{fig:feature_sampling}. We observe that with shared G-MLP, both approaches outperform the baseline ImVoxelNet, and sampling from image features yields better performance ($+0.7$ in mAP@0.25) than sampling from volume features. 
% sampling features from 3D volume still improves the performance (1.9 mAP@.25 compared to ImVoxelNet), yet the performance gain is not as large as sampling features onto ray from multi-view 2D features.
For the novel-view synthesis task using NeRF branch, sampling from image features achieves $20.51$ in PSNR comparing to $18.93$ with volume sampling.

The fact that the performance of the NeRF branch is proportional to the performance of the detection branch also indicates that better NeRF optimization could lead to better detection results. 
% This indicates better optimizing NeRF also benefits detection. 

\begin{table}
\centering
\captionsetup{aboveskip=2pt}\captionsetup{belowskip=0pt}\captionof{table}{Ablation study for loss.}
\label{tab:ablation_loss}
\footnotesize
\resizebox{0.42\textwidth}{!}{
\begin{tabular}{c|c|c|c}
% @{\hspace{1.0\tabcolsep}}c@{\hspace{1.0\tabcolsep}}c@{\hspace{1.0\tabcolsep}}c@{\hspace{1.0\tabcolsep}}c} 
% \begin{tabular}{ p{25mm}<{\centering}| p{10mm}<{\centering} | p{12mm}<{\centering}  p{10mm}<{\centering}  p{10mm}<{\centering} p{10mm}<{\centering} p{10mm}<{\centering} p{12mm}<{\centering} | p{10mm}<{\centering}}

\toprule
Photo-metric loss &  Depth loss & mAP@.25  & mAP@.50 \\
\midrule
% \multirowcell{4}{NeRF-Det-R50-1x}  
$\checkmark$ & $\checkmark$ &  50.1 & 24.4 \\
$\checkmark$ & - &  49.4 & 24.1 \\
 - & $\checkmark$ &   50.0 & 24.2 \\
 - & - &  48.5 & 23.6\\
\bottomrule
\end{tabular}}
\vspace{-0.35cm}
\end{table}

\vspace{-0.5cm}
\paragraph{Ablation study on different loss.} We study how different losses work for the NeRF branch, as shown in Tab.~\ref{tab:ablation_loss}. It shows that with only photo-metric loss, the performance is closed to purely using depth supervision (third row) in term of mAP@.50, indicating that the multi-view RGB consistency already provides sufficient geometry cues to enable the NeRF branch to learn geometry. When using both photo-metric loss and depth loss, the performance can be further improved. When neither photo-metric loss nor depth loss is used (last row), the performance falls back to that of a cost-volume based method. The performance is dropped by 1.2 mAP@.25 and 0.5 mAP@.50, which demonstrates that the NeRF branch is more effective.

\begin{table}
\centering
\captionsetup{aboveskip=2pt}\captionsetup{belowskip=0pt}\captionof{table}{Ablation study on augmented features.}
\label{tab:ablation_feature}
\footnotesize
\resizebox{0.38\textwidth}{!}{
\vspace{-0.1cm}
\begin{tabular}{c|c|c|c|c}
% @{\hspace{1.0\tabcolsep}}c@{\hspace{1.0\tabcolsep}}c@{\hspace{1.0\tabcolsep}}c@{\hspace{1.0\tabcolsep}}c} 
% \begin{tabular}{ p{25mm}<{\centering}| p{10mm}<{\centering} | p{12mm}<{\centering}  p{10mm}<{\centering}  p{10mm}<{\centering} p{10mm}<{\centering} p{10mm}<{\centering} p{12mm}<{\centering} | p{10mm}<{\centering}}

\toprule

Avg &  Var  & Color & mAP@.25 & mAP@.50 \\
\midrule
 
$\checkmark$ & $\checkmark$ &  $\checkmark$ &  50.1 & 24.4 \\ $\checkmark$ & $\checkmark$ &  -  &  49.7 & 24.3 \\
$\checkmark$ & -            &  -  &  49.0& 23.7 \\
\bottomrule
\end{tabular}}
%\vspace{-0.5cm}

\end{table}

\vspace{-0.5cm}
\paragraph{Ablation study on different features.} We then study how different features affect performance, as shown in Tab. \ref{tab:ablation_feature}. The experiment shows that introducing variance features improves performance significantly -- over 0.7 mAP@.25 and 0.6 mAP@.50 compared to using only average features, which demonstrates that variance features indeed provide a good geometry prior. Moreover, incorporating image features also improves performance, indicating that low-level color information also provides good geometry cues.

\begin{table}
\centering
\captionsetup{aboveskip=2pt}\captionsetup{belowskip=0pt}\captionof{table}{Ablation on how detection branch influences novel view synthesis (NVS) and depth estimation (DE) on test set.}
\label{tab:nerf_evaluate}
\footnotesize
\resizebox{0.47\textwidth}{!}{
\begin{tabular}{l|c|c|c}
% @{\hspace{1.0\tabcolsep}}c@{\hspace{1.0\tabcolsep}}c@{\hspace{1.0\tabcolsep}}c@{\hspace{1.0\tabcolsep}}c} 
% \begin{tabular}{ p{25mm}<{\centering}| p{10mm}<{\centering} | p{12mm}<{\centering}  p{10mm}<{\centering}  p{10mm}<{\centering} p{10mm}<{\centering} p{10mm}<{\centering} p{12mm}<{\centering} | p{10mm}<{\centering}}

\toprule

Method & PSNR (NVS) $\uparrow$ & SSIM (NVS) $\uparrow$ & RMSE (DE) $\downarrow$\\
\midrule
w/ Det branch & 20.51 &  0.83 & 0.756 \\
w/o Det branch & 20.94 & 0.84 & 0.747\\
% w/o Det Branch and Depth loss & & \\

\bottomrule
\end{tabular}}
\vspace{-0.5cm}
\end{table}
\vspace{-0.5cm}
\paragraph{Ablation Study on detection branch affecting novel view synthesis.} We keep the target views and source views the same with and without the detection branch. Results are shown in Tab.~\ref{tab:nerf_evaluate}. While the NeRF branch significantly improves 3D detection by improving scene geometry modeling, the detection branch does not benefit the NeRF branch. In fact, disabling the detection branch brings a 0.43db improvement. We hypothesize that the detection branch is prone to erase low-level details which are needed for the NeRF, which we aim to address in our future work.

\section{Conclusion}
\vspace{-0.2cm}
In this paper, we present NeRF-Det, a novel method that uses NeRF to learn geometry-aware volumetric representations for 3D detection from posed RGB images. We deeply integrate multi-view geometry constraints from the NeRF branch into 3D detection through a subtle shared geometry MLP. To avoid the large overhead of per-scene optimization, we propose leveraging augmented image features as priors to enhance the generalizablity of NeRF-MLP. In addition, we sample features from high resolution images instead of volumes to address the need for high-resolution images in NeRF. Our extensive experiments on the ScanNet and ARKITScene datasets demonstrate the effectiveness of our approach, achieving state-of-the-art performance for indoor 3D detection using RGB inputs. Notably, we observe that the NeRF branch also generalizes well to unseen scenes. Furthermore, our results highlight the importance of NeRF for 3D detection, and provide insights into the key factors for achieving optimal performance in this direction.

\section{Acknowledgment}
\noindent We sincerely thank Chaojian Li for the great help on NeuralRecon experiments, Benran Hu for the valuable advice on the NeRF-RPN experiments, Feng (Jeff) Liang and Hang Gao for the insightful discussions, as well as Matthew Yu and Jinhyung Park for the paper \mbox{proofreading}. 
% \clearpage
{\small
\bibliographystyle{ieee_fullname}
\bibliography{egbib}
}
\appendix

\section{Dataset and Implementation Details}
\paragraph{Dataset.} Our experiments are conducted on ScanNetV2 \cite{dai2017scannet} and ARKITScenes dataset \cite{arkitscenes}. ScanNetV2 dataset is a challenging dataset containing 1513 complex scenes with around 2.5 million RGB-D frames and annotated with semantic and instance segmentation for 18 object categories. Since ScanNetV2 does not provide amodal or oriented bounding box annotation, we predict axis-aligned bounding boxes instead, as in \cite{hou20193d,qi2019deep,rukhovich2022imvoxelnet}. We mainly evaluate the methods by mAP with 0.25 IoU and 0.5 IoU threshold, denoted by mAP@.25 and mAP@.50. 

ARKITScenes dataset contains around 1.6K rooms with more than 5000 scans. Each scan includes a series of RGB-D posed images. In our experiments, we utilize the subset of the dataset with low-resolution images. The subset contains 2,257 scans of 841 unique scenes, and each image in the scan is of size $256\times192$. We follow the dataset setting provided by the official repository \footnote{https://github.com/apple/ARKitScenes/tree/main/threedod}. We mainly evaluate the methods by mAP with 0.25 IoU as follow \cite{arkitscenes}.

\paragraph{Detection Branch.} We follow ImVoxelNet, mainly use ResNet50 with FPN as our backbone and the detection head consists of three 3D convolution layers for classification, location, and centerness, respectively. For the experiment on the ARKITScenes, we additionally predict the rotation. We use the same size $40 \times 40 \times 16$ of the voxels, with each voxel represents a cube of $0.16m, 0.16m, 0.2m$. Besides, we also keep the training recipe as same as ImVoxelNet. During training, we use 20 images on the ScanNet datatset and 50 images on the ARKITScenes dataset by default. During test we use 50 images and 100 images on the ScanNet dataset and ARKITScenes dataset, respectively. The network is optimized by Adam optimizer with an initial learning rate set to 0.0002 and weight decay of 0.0001, and it is trained for 12 epochs, and the learning rate is reduced by ten times after the $8\-th$ and $11\-th$ epoch.

\paragraph{NeRF Branch.} In our NeRF branch, 2048 rays are randomly sampled at each iteration from 10 novel views for supervision. Note that the 10 novel views are ensured to be different with the views input to detection branch for both training and inference. We set the near-far range as (0.2 meter - 8 meter), and uniformly sample 64 points along each ray. During volumetric rendering, if more than eight points on the ray are projected to empty space, then we would throw it and do not calculate the loss of the ray. The geometry-MLP (G-MLP) is a 4-layer MLP with 256 hidden units and skip connections. The color-MLP (C-MLP) is a one-layer MLP with 256 hidden units. Our experiments are conducted on eight V100 GPUs with 16G memory per GPU. We batched the data in a way such that each GPU carries a single scene during training. During training, the two branches are end-to-end jointly trained. During inference, we can keep either one of the two branches for desired tasks. The whole Our implementation is based on MMDetection3D~\cite{mmdet3d2020}. 

\section{Evaluation Protocol of Novel View Synthesis and Depth Estimation.} To evaluate the novel view synthesis and depth estimation performance, we random select 10 views of each scene as the novel view (indicated as target view in IBRNet \cite{wang2021ibrnet}), and choose the nearby 50 views as the support views. To render the RGB and depth for the 10 novel views, each points shooting from the pixels of novel views would be projected to the all support views to sample features, and then pass into the NeRF MLP as illustrated in Method section. We keep the same novel view and support view for both setting in Table. 6 of the main text. Note that the evaluation is conducted on the test set of ScanNet dataset, which are never seen during training. The non-trivial results also demonstrate the generazability of the proposed geometry-aware volumetric representation.

\begin{figure*}[t!]
	\begin{center}
% 		\vspace{-0.4cm}
	\includegraphics[width=1.0\linewidth]{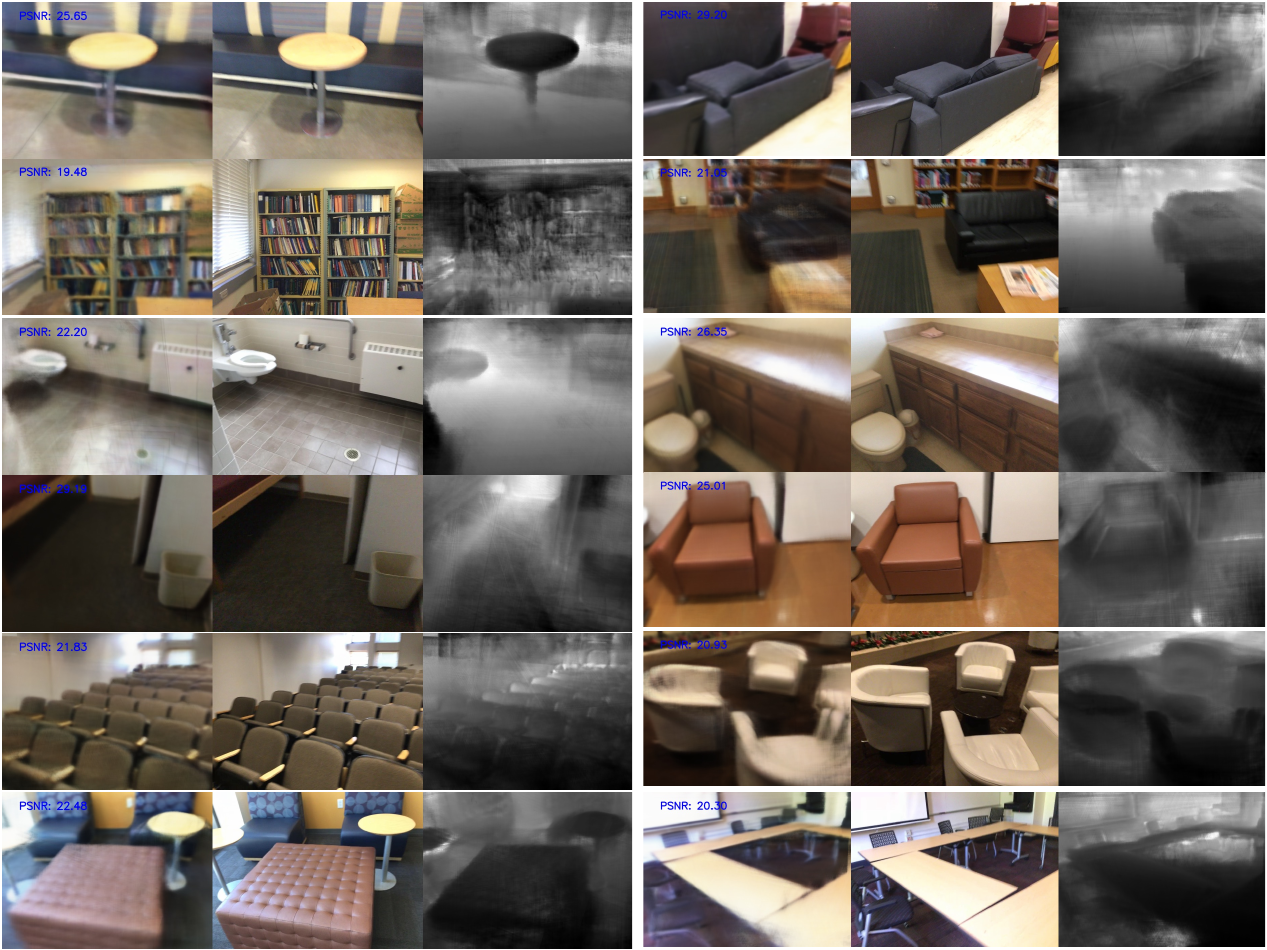}
		\captionof{figure}{
		Novel-view synthesis results on top of NeRF-Det-R50-2x*. For each triplet group, the left figure is the synthesized results, the middle one is the ground truth RGB image, and the right part is the estimated depth map. Note that the visualization is from test set, which is never seen during training.
		}
		\label{fig:nerf_spp}
	\end{center}
\end{figure*}

\section{Additional Results}
\noindent\textbf{Ablation studies on number of views.}
\begin{table}
\centering
\captionsetup{aboveskip=2pt}\captionsetup{belowskip=0pt}\captionof{table}{Ablation on number of views. Due to the GPU memory limitation, we downsample the image resolution 2x when conduct experiments on 100 views (denoted as ImVoxelNet-R50-2x' and NeRF-Det-R50-2x'.). Experiments on each setting run three times. We report the mean and standard deviations of our experiments. }
\label{tab:number_of_views}
\footnotesize
\resizebox{.5\textwidth}{!}{
\begin{tabular}{l|c|c}
% @{\hspace{1.0\tabcolsep}}c@{\hspace{1.0\tabcolsep}}c@{\hspace{1.0\tabcolsep}}c@{\hspace{1.0\tabcolsep}}c} 
% \begin{tabular}{ p{25mm}<{\centering}| p{10mm}<{\centering} | p{12mm}<{\centering}  p{10mm}<{\centering}  p{10mm}<{\centering} p{10mm}<{\centering} p{10mm}<{\centering} p{12mm}<{\centering} | p{10mm}<{\centering}}

\toprule
\textbf{Methods} & mAP@.25 & mAP@.50 \\
\midrule
% ImVoxelNet-R50-1x (baseline) & 47.5 & 23.3 \\
% GT-Depth-R50-1x (upper-bound) & 51.9 \textcolor{darkgreen}{(+4.4)} & 27.0 \textcolor{darkgreen}{(+3.7)}\\
% NeuralRecon-Depth-R50-1x & 47.8 \textcolor{darkgreen}{(+0.4)} & 20.1 \textcolor{red}{(-3.2)}\\
% Cost-Volume-R50-1x (sigmoid) & 48.5 \textcolor{darkgreen}{(+1.0)} & 23.6 \textcolor{darkgreen}{(+0.3)} \\
% \rowcolor[gray]{.9} 
% NeRF-Det-R50-1x (w/o NeRF branch) & 48.2 \textcolor{darkgreen}{(+0.7)} & 23.6 \textcolor{darkgreen}{(+0.3)} \\
% \rowcolor[gray]{.9} 
% NeRF-Det-R50-1x & 49.5 \textcolor{darkgreen}{(+2.0)} & 24.1 \textcolor{darkgreen}{(+0.8)} \\ % Us!
% \rowcolor[gray]{.9} 
% NeRF-Det-R50-1x* & 50.1 \textcolor{darkgreen}{(+2.6)} & 24.4 \textcolor{darkgreen}{(+1.1)}\\
% \midrule
ImVoxelNet-R50-2x (10 views) & 37.8$\pm$1.2 & 17.5$\pm$1.0  \\
ImVoxelNet-R50-2x (20 views) & 46.5$\pm$0.5 & 21.1$\pm$0.5  \\
ImVoxelNet-R50-2x (50 views) & 48.4$\pm$0.3 & 23.7$\pm$0.2  \\
ImVoxelNet-R50-2x'(100 views)& 48.1$\pm$0.1& 24.7$\pm$0.1 \\

\rowcolor[gray]{.9} 
NeRF-Det-R50-2x (10 views)   & 41.4 $\pm$1.0 \textcolor{darkgreen}{(+3.6)} & 19.2$\pm$0.9 \textcolor{darkgreen}{(+1.7)}  \\
\rowcolor[gray]{.9} 
NeRF-Det-R50-2x (20 views)   &  50.2 $\pm$0.5 \textcolor{darkgreen}{(+3.7)}&  23.6$\pm$0.4 \textcolor{darkgreen}{(+2.5)}\\ % Us!
\rowcolor[gray]{.9} 
NeRF-Det-R50-2x (50 views)   &  51.8 $\pm$0.2 \textcolor{darkgreen}{(+3.4)}&  26.0$\pm$0.1 \textcolor{darkgreen}{(+2.3)}\\
\rowcolor[gray]{.9} 
NeRF-Det-R50-2x'(100 views) & 52.2$\pm$0.1  \textcolor{darkgreen}{(+4.1)} & 27.4$\pm$0.1 \textcolor{darkgreen}{(+2.7)} \\

% ImVoxelNet (1x) & 47.5 & 23.3 \\
% ImVoxelNet (2x) & 48.4 & 24.4 \\
% ImVoxelNet (4x) & 48.9 & 25.1 \\
% ImVoxelNet (1x w/ depth from NeuralRecon) & 47.8 & 20.1\\
% % ImVoxelNet (1x w/ depth from GT) & 51.9 & 27.0 \\
% \rowcolor[gray]{.9} 
% NeRF-Det (1x) & 49.4 & 24.1 \\ % Us!
% \rowcolor[gray]{.9} 
% % NeRF-Det (2x)  & 52.0 & 26.1 \\ % Us!
% % \rowcolor[gray]{.9} 
% % NeRF-Det (4x) & 51.6  & 27.5 \\ % Us!
% \rowcolor[gray]{.9} 
% NeRF-Det (1x w/ depth supervision) & 50.1 & 24.4 \\ % Us!

\bottomrule
\end{tabular}}
\vspace{-0.55cm}
\end{table}
We conducted an analysis of how the number of views affects the performance of 3D detection, as shown in Table~\ref{tab:number_of_views}. Specifically, we used the same number of training images (20 images) and tested with different numbers of images. Our proposed NeRF-Det-R50-2x showed a significant improvement in performance as the number of views increased. In contrast, the performance of ImVoxelNet-R50-2x had limited improvement, and even worse, the performance decreased when the number of views increased to 100. We attribute the performance improvements of NeRF-Det to its effective scene modeling. NeRF performs better as the number of views increases, typically requiring over 100 views for an object \cite{mildenhall2021nerf}. Our proposed NeRF-Det inherits this advantage, leading to a drastic performance gain of 4.1 mAP@.25 and 2.7 mAP@.50 on 100 views.

Overall, our analysis demonstrates the effectiveness of our proposed NeRF-Det in leveraging multi-view observations for 3D detection and the importance of utilizing a method that can effectively model the scene geometry.

\noindent\textbf{More Qualitative Results}
We provide more visualization results of novel-view synthesis and depth estimation, as shown in Fig. \ref{fig:nerf_spp}. The results come from the test set of ScanNet. We can observe that the proposed method generalizes well on the test scenes. Remarkably, it achieves non-trivial results on the relatively hard cases. For example, the left of the second row presents a bookshelf with full of colorful books, our method can give reasonable novel-view synthesis results. On the other hand, for the left of fifth row, the extremely dense chairs are arranged in the scenes and we can observe the method can predict accurate geometry.

\section{Discussion about outdoor 3D detection}
We emphasize the differences of NeRF-Det and the other 3D detection works in outdoor scenes. Our proposed NeRF-Det shares the similar intuition with many of outdoor 3D detection works, such as \cite{park2022time,wang2022monocular,li2022bevstereo}, which try to learn geometric-aware representations. However, the proposed NeRF-Det and the other works differ intrinsically. The outdoor 3D detection works \cite{park2022time,wang2022monocular,li2022bevstereo} propose to use cost volume or explicitly predicted depth to model the scene geometry. Instead, NeRF-Det leverage the discrepancy of multi-view observations, i.e., the augmented variance features in our method section, as the priors of NeRF-MLP input. Beyond the cost volume, we step forward to leverage the photo-realistic principle to predict the density fields, and then transform it into the opacity field. Such a geometry representation is novel to the 3D detection task. The analysis in our experiment part also demonstrates the advantages of the proposed opacity field. In addition to the different method of modeling scene geometry, our design of combining NeRF and 3D detection in an end-to-end manner allows the gradient of NeRF to back-propagate and benefit the 3D detection branch. This is also different from previous NeRF-then-perception works \cite{hu2022nerf,vora2021nesf}.

Our NeRF-Det is specifically designed for 3D detection in indoor scenes, where objects are mostly static. Outdoor scenes present unique challenges, including difficulties in ensuring multi-view consistency due to moving objects, unbounded scene volume, and rapidly changing light conditions that may affect the accuracy of the RGB value used to guide NeRF learning. We plan to address these issues and apply NeRF-Det to outdoor 3D detection in future work.

\end{document}